\documentclass{article}
\usepackage{ijcai21}

\usepackage{times}
\usepackage{soul}
\usepackage{url}
\usepackage[hidelinks]{hyperref}
\usepackage[utf8]{inputenc}
\usepackage[small]{caption}
\usepackage{graphicx}
\usepackage{amsmath}
\usepackage{amsfonts}
\usepackage{amssymb}
\usepackage{amsthm}
\usepackage{booktabs}
\usepackage{algorithm}
\usepackage{algorithmic}
\usepackage{color}
\usepackage{threeparttable}
\usepackage{enumitem}
\usepackage{multirow}
\usepackage{makecell}
\usepackage{bm}
\usepackage{subfigure}
\usepackage{caption}

\title{Domain-Smoothing Network for Zero-Shot Sketch-Based Image Retrieval}

\author{
Zhipeng Wang\and
Hao Wang\and
Jiexi Yan\and
Aming Wu$^*$\And
Cheng Deng\footnote{Contact Author}\\
\affiliations
Xidian University\\
\emails
\{zpwang1996, haowang.xidian\}@gmail.com,
jxyan@stu.xidian.edu.cn,
amwu@xidian.edu.cn,
chdeng.xd@gmail.com
}

\begin{document}

\maketitle

\begin{abstract}
Zero-Shot Sketch-Based Image Retrieval (ZS-SBIR) is a novel cross-modal retrieval task, where abstract sketches are used as queries to retrieve natural images under zero-shot scenario. Most existing methods regard ZS-SBIR as a traditional classification problem and employ a cross-entropy or triplet-based loss to achieve retrieval, which neglect the problems of the domain gap between sketches and natural images and the large intra-class diversity in sketches. Toward this end, we propose a novel Domain-Smoothing Network (DSN) for ZS-SBIR. Specifically, a cross-modal contrastive method is proposed to learn generalized representations to smooth the domain gap by mining relations with additional augmented samples. Furthermore, a category-specific memory bank with sketch features is explored to reduce intra-class diversity in the sketch domain. Extensive experiments demonstrate that our approach notably outperforms the state-of-the-art methods in both Sketchy and TU-Berlin datasets. Our source code is publicly available at \url{https://github.com/haowang1992/DSN}.
\end{abstract}

\section{Introduction}
Sketch-based image retrieval (SBIR) \cite{sangkloy2016sketchy,bhunia2020sketch} is an interesting and practical task aiming to retrieve similar images based on the queries of the hand-drawn sketches. With the help of large-scale labeled datasets, SBIR can achieve astonishing performance. However, collecting and annotating large-scale datasets is time-consuming and labor-intensive. To this end, zero-shot SBIR (ZS-SBIR) \cite{shen2018zero} is proposed, which aims to retrieve categories or samples unseen at the training stage. However, solving ZS-SBIR is significantly challenging since it simultaneously deals with the inherent domain gap and limited knowledge about the unseen categories \cite{deng2020progressive}. 

\begin{figure}[t]
    \begin{center}
    \includegraphics[width=0.9\linewidth]{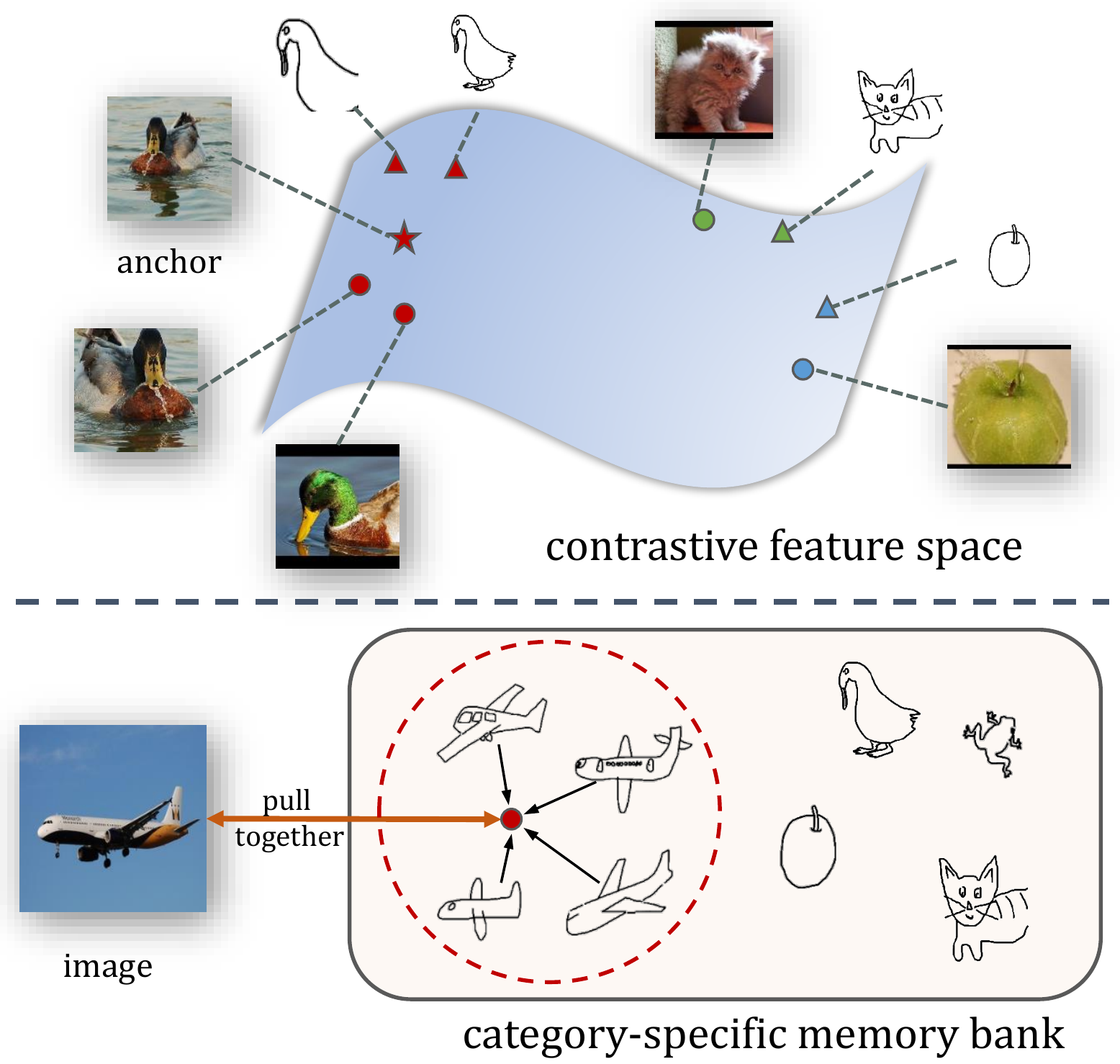}
    \caption{The illustration of our DSN method. (Top) To smooth the domain gap between sketches and images, we propose a cross-modal contrastive method to mine the relations between samples with sufficiently positive and negative samples. (Bottom) Besides, a category-specific memory bank is built to address the large intra-class diversity in the sketch domain, where prototypes of each sketch category are learned to reduce the variability in appearance.}\label{fig:intro}
    \end{center}
 \end{figure}

Previous ZS-SBIR methods \cite{xu2020progressive} attempt to learn a common representation of sketches and images with cross-entropy loss or triplet-based loss, which neglect the impact of the inherent domain gap. Specifically, cross-entropy loss equally treats sketches and images for classification and neglects the relations between sketches and images of the same category.  Besides, triplet-based loss is usually utilized to measure the relations between samples. However, due to the limited positive and negative samples, triplet-based loss may not mine the relations sufficiently. Furthermore, sketches within the same category might be extremely diverse in appearance due to the drawing styles of different people, which is ignored by most methods and degrades retrieval performance. 

To alleviate the domain gap and the intra-class diversity in sketch domain, we propose a Domain-Smoothing Network (DSN) for ZS-SBIR (seen Fig.~\ref{fig:intro}), mainly including a cross-modal contrastive method and a category-specific memory bank. Specifically, a pre-trained CNN backbone is first applied to extract visual features of the images and sketches. To smooth the domain gap between the sketches and images, a cross-modal contrastive method is then conducted after a projection head for visual features. Meanwhile, a category-specific memory bank is proposed, where sketch category prototypes are learned to reduce the intra-class diversity in the sketch domain. Besides, to enhance the discrimination of obtained representations, our model is trained with the knowledge of the teacher model and label information inspired by previous work on teacher signal \cite{liu2019semantic}. To validate the superiority of our DSN method, extensive experiments are conducted and the results demonstrate our method remarkably outperforms the state-of-the-art methods on two popular datasets.

The main contributions of this work are summarized as:
\begin{itemize}
\item We propose the DSN model for ZS-SBIR task, where a cross-modal contrastive method is utilized to mine the relations between sketches and images with sufficient positive and negative samples to significantly smooth the domain gap.
\item To address the large intra-class diversity in the sketch domain, we elaborate a category-specific memory bank to learn prototypes for each sketch category to reduce the variability in appearance.
\item Extensive experiments conducted on two popular datasets, {\it i.e.}, Sketchy and TU-Berlin, demonstrate that our approach outperforms the state-of-the-art methods by a large margin.
\end{itemize}

\begin{figure*}[t]
    \begin{center}
        \includegraphics[width=0.9\linewidth]{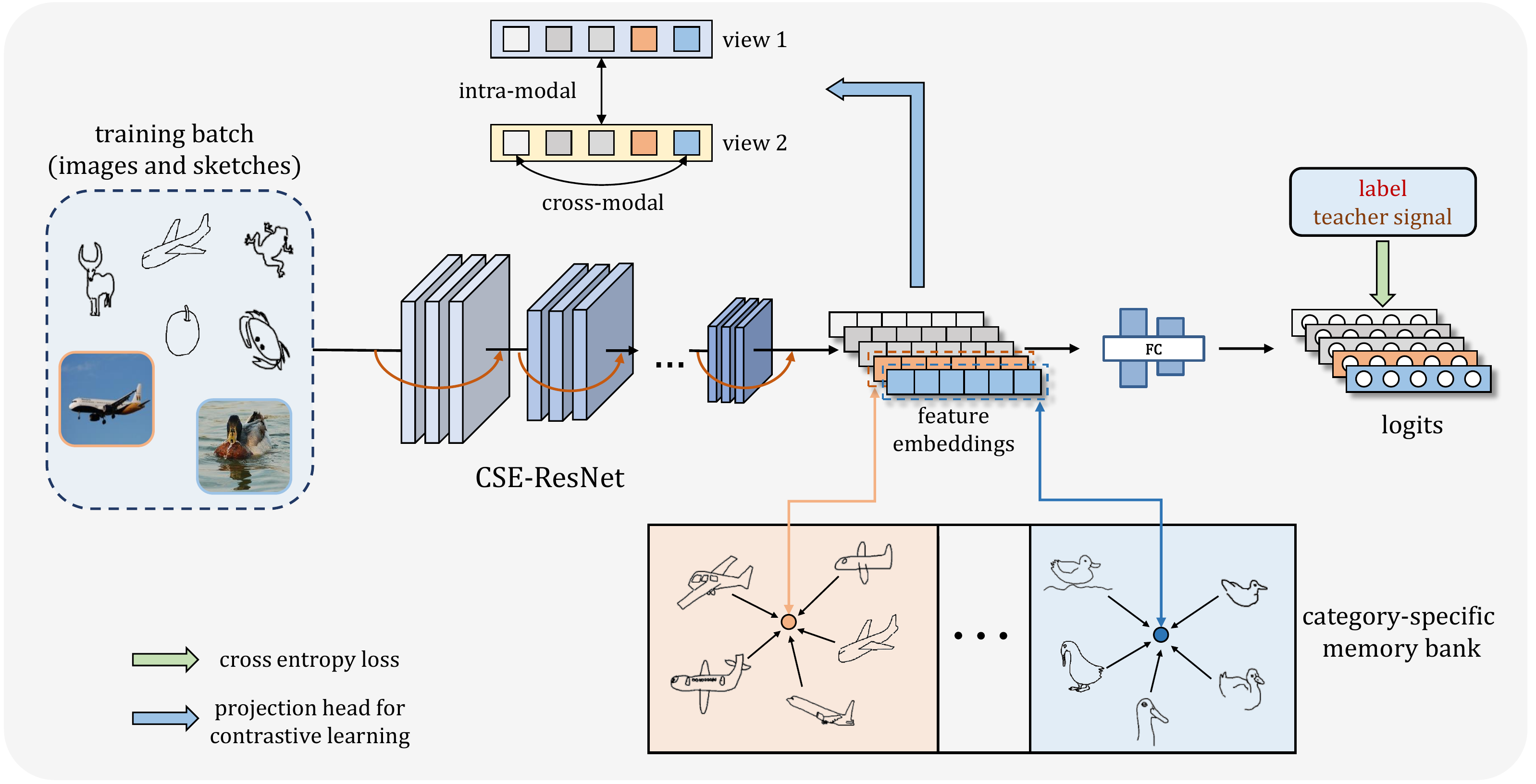}
    \caption{An overview of our model. We use a CSE-ResNet-50 as the backbone to learn a common representation for input sketches and images. After obtaining feature embeddings, three branches of modules are applied. The first one is to employ cross-modal contrastive method in different views after a projection head with fully-connected layers. The second one is to enforce the image feature to be closer to the prototype of sketch features in the category-specific memory bank. The last one is to leverage the teacher signal and label information to train the whole model.}\label{fig:framework}
    \end{center}
 \end{figure*}

\section{Related Work}
Since ZS-SBIR is an extremely challenging task integrated with SBIR and zero-shot learning (ZSL), we will review both of them in this section.
\subsection{Sketch-Based Image Retrieval}
The main target of SBIR task is to alleviate the domain gap between image and sketch by aligning their features in a common semantic space. Early works utilize hand-crafted features to represent sketches and match them with the edge maps of images by various Bag-of-Words model. However, it is difficult to align the edge map with hand-crafted features directly, which makes the problem of domain gap unresolved. With the advent of deep neural networks, researchers tend to employ CNN to extract deep features from images and sketches to learn transferrable representation. Among them, Siamese architecture based methods \cite{qi2016sketch} play an important role by handling sketches and images separately with ranking loss. Compared with these methods, our DSN model proposes a cross-modal contrastive method to address the domain gap effectively. By pulling together samples with the same label and pushing apart those from different categories in different modalities and views, the cross-modal contrastive method can mine the relations between sketches and images sufficiently to smooth the inherent domain gap.

\subsection{Zero-Shot Learning}


Zero-shot learning (ZSL) aims to recognize objects whose instances are unseen during the training period \cite{xian2018zero,wei2019adversarial,wei2020lifelong}. Existing ZSL methods can be classified into two categories: embedding-based \cite{wei2020incremental} and generative-based \cite{li2021generalized} approaches. For the embedding-based method, \cite{kodirov2017semantic} learned an auto-encoder to map visual features to semantic embeddings, where the nearest neighbor search is conducted. For the generative-based method,  \cite{xian2018feature} proposed a feature generating network to synthesize unseen visual features to mitigate the data imbalance problem in ZSL. Besides, various semantic representation methods have been proposed, including attributes annotations \cite{akata2015label} and word vectors \cite{mikolov2013efficient} to bridge the seen and unseen categories. Due to the space limitation, more details of ZSL can refer to \cite{xian2018zero}.

\subsection{Zero-Shot Sketch-Based Image Retrieval}
ZS-SBIR is a novel task that conducts SBIR under zero-shot setting and is more practical in realistic applications. \cite{shen2018zero} first combined ZSL with SBIR, and proposed a cross-modal hashing method for ZS-SBIR. \cite{yelamarthi2018zero} built an auto-encoder to generate image features by sketches for alignment, while \cite{dutta2019semantically} proposed a paired cycle-consistent generative model with adversarial training to align the sketch and image in the same semantic space. \cite{dutta2019style} presented a feature disentanglement method, where the images generated by sketches and corresponding styles can be conducted in retrieval as queries. \cite{liu2019semantic} treated ZS-SBIR as a catastrophic forgetting problem and employed a CNN model pre-trained on ImageNet as the teacher model to preserve the knowledge. However, all these methods ignore the large intra-class diversity in the sketch domain. In this work, we propose a category-specific memory bank to learn prototypes for each sketch category to reduce the intra-class diversity in the sketch domain.

\section{Methodology}

\subsection{Problem Definition}
In ZS-SBIR, the dataset can be divided into two parts: the training set and the testing set. The training set, denoted as $\mathcal{D}^{seen}=\{ \mathcal{I}^{seen}, \mathcal{S}^{seen}\}$, is utilized for training the retrieval model, where $I^{seen}$ and $S^{seen}$ represent images and sketches from seen categories, respectively. Similarly, the testing set is utilized for validating the performance of the retrieval model, which can be denoted as $\mathcal{D}^{unseen}=\{ \mathcal{I}^{unseen}, \mathcal{S}^{unseen}\}$.  Mathematically, let $\mathcal{I}^{seen}=\{ ( I_j, y_j) | y_j \in \mathcal{C}^{seen}\}^{N_1}_{j=1}$ and $\mathcal{S}^{seen}=\{ ( S_i, y_i) | y_i \in \mathcal{C}^{seen}\}^{N_2}_{i=1}$, where $y$ represents the category label. $\mathcal{C}^{seen}$ denotes the seen category set. $N_1$ and $N_2$ denote the number of images and sketches in training set, respectively.

During the training period, we sample batches of images and sketches from the training set, where corresponding labels are utilized to guide the model to learn discriminative features. And at testing stage, a sketch query $S' \in \mathcal{S}^{unseen}$ with its label $y \in \mathcal{C}^{unseen}$, where $\mathcal{C}^{unseen}$ denotes the unseen category set, is given to match the similar image $I' \in \mathcal{I}^{unseen}$. If the label $y'$ of the retrieved image is the same as that of the given sketch, {\it i.e.}, $y=y'$, we name this as a successful retrieval. In zero-shot setting, there is no testing category appears in training period, {\it i.e.}, $\mathcal{C}^{seen} \bigcap \mathcal{C}^{unseen}=\varnothing $.

\subsection{Domain-Smoothing Network}
The architecture of our DSN model is illustrated in Figure \ref{fig:framework}. Specifically, we employ a CSE-ResNet-50 backbone (CSE) \cite{lu2018learning} to extract feature embeddings for images and sketches. To smooth the domain gap between sketch and image, we propose a cross-modal contrastive method to align these two domains. Besides, a category-specific memory bank is built to address the large intra-class diversity in sketch domain. The details of these modules will be discussed in this section.

\subsubsection{Cross-modal Contrastive Method}
The domain gap is an inherent challenge in SBIR task. Previous methods usually use a unified framework to learn a common representation of sketches and images by cross-entropy loss or triplet-based loss. However, cross-entropy loss may neglect the relations between sketches and images as mentioned above. Besides, triplet-based loss is usually utilized to measure the relations between samples. However, due to the limited positive and negative samples, triplet-based loss may not mine the relations sufficiently.

To effectively smooth the domain gap, we propose a cross-modal contrastive method. In detail, for each image $I$ and sketch $S$ from $\mathcal{D}^{seen}$, two random augmentations are generated for contrastive learning by augmentation method \cite{khosla2020supervised}, denoted as $\{ \hat{I},\tilde{I},\hat{S},\tilde{S} \}$. Then CSE is applied to extract feature embeddings for each image, sketch and these augmentations as:
\begin{equation}
\begin{split}
[f_{img}, \hat{f}_{img}, \tilde{f}_{img}] =& \; [\text{CSE}(I), \text{CSE}(\hat{I}),\text{CSE}(\tilde{I})], \\
[f_{ske}, \hat{f}_{ske}, \tilde{f}_{ske}] =& \; [\text{CSE}(S), \text{CSE}(\hat{S}),\text{CSE}(\tilde{S})]. 
\end{split}
\end{equation}
After obtaining feature embeddings $f$, a projection head with two fully-connected layers, named as $P$, is utilized for mapping $f$ to vector $v=P(f)$. In this way, we can obtain the vectors for two augmentations of sketch and image, which can be written as $\{ \hat{v}_{img}, \tilde{v}_{img}, \hat{v}_{ske}, \tilde{v}_{ske}\} \subseteq V$. To illustrate clearly, we denote $\hat{v}_{img}$ as the anchor. It should be noted that even though $\hat{v}_{img}$ and $\tilde{v}_{img}$ are in different views, they share the same label. So $\tilde{v}_{img}$ is an {\bf intra-modal} positive sample for $\hat{v}_{img}$ . Similarly, $\hat{v}_{ske}$ and $\tilde{v}_{ske}$ are {\bf cross-modal} samples for $\hat{v}_{img}$. Their roles of positive or negative depends on whether their labels are the same as $\hat{v}_{img}$. Then contrastive method is conducted to pull together the anchor with the positive samples and push apart the anchor from the negative samples in a cross-modal scenario.

Our cross-modal contrastive method (CMCM) loss can be summarized as:
\begin{equation}
    \resizebox{.91\linewidth}{!}{$
        \displaystyle
        \mathcal{L}_{cmcm}=\sum_{i=1}^{|V|}\left[ \frac{-1}{2N_i-1} \sum_{j = 1}^{2|V|}\bm{1}_{y_i=y_j}\log \frac{\exp (v_i\cdot v_j / \tau )}{ {\textstyle \sum_{k=1}^{2|V|}} \bm{1}_{i \neq k}\exp (v_i\cdot v_k/ \tau)} \right],   
    $}
\end{equation}%
where for each vector $v \in V$, there is a corresponding label $y$. $|V|$ represents the cardinality of the vector set $V$, and $N_i=|\{ y_j | y_j=y_i, j \neq i, v_i \in V, v_j \in V\}|$. $\bm{1}_{\text{condition}} \in \{ 0,1\}$ is an indicator function evaluating to 1 iff condition is true. $\tau$ is the temperature parameter for contrastive learning.

\subsubsection{Category-specific Memory Bank}
To reduce the large intra-class diversity in the sketch domain, we propose a category-specific memory bank to learn a prototype representation for sketches from the same category, where the image is enforced to be closer to the prototype of top-$k$ similar sketches at the feature level. 

Specifically, for each training batch, we can obtain a batch of feature embeddings $\{ \mathcal{F}_{img}, \mathcal{F}_{ske}\}$, where $\mathcal{F}_{img}=\{f_{img} | y_{img} \in \mathcal{C}_{seen}\}$ and $\mathcal{F}_{ske}=\{f_{ske}| y_{ske} \in \mathcal{C}_{seen}\}$. In order to update the memory bank and leverage the history knowledge at the same time, we maintain a stack filled with $k$ sketch features for each category adaptively. First, for $f_{ske}$ and its label $y_{ske}$, we search for the corresponding image features $\mathcal{F}'_{img}$ with the same label in a batch, where $\mathcal{F}'_{img}=\{ f_{img} | y_{img} = y_{ske}\}$. The mean value of corresponding image features $f'_{img}$ can be obtained as:
\begin{equation}
    f'_{img}=\frac{1}{|\mathcal{F}'_{img}|} \sum_{f_{img} \in \mathcal{F}'_{img}} f_{img}. 
\end{equation}
Next, we calculate the cosine similarity between $f'_{img}$ and $\mathcal{F}'_{ske} \cup \{ f_{ske}\}$ to select new top-$k$ similar sketch features and store them in the memory bank, where $\mathcal{F}'_{ske}$ is the previous top-$k$ sketch features.

After updating the memory bank, the prototype of new top-$k$ sketch features can be obtained as:
\begin{equation}
    p_{ske}=\frac{1}{|\mathcal{M}_{ske}|} \sum_{f_{ske} \in \mathcal{M}_{ske}} f_{ske},
\end{equation}
where $\mathcal{M}_{ske}$ denotes the new top-$k$ sketch features. Then the memory loss can be formulated as:
\begin{equation}
    \mathcal{L}_{ml}=-\frac{f_{img}^\top  \cdot p_{ske}}{|f_{img}| \cdot |p_{ske}|},
\end{equation}
where $|\cdot |$ indicates the norm of the feature vector. 

\subsubsection{Discriminative Learning}
To learn discriminative representations, we use cross-entropy loss to align the learned feature with its label:
\begin{equation}
    \mathcal{L}_{cls}=-\sum_{i = 1}^{N}\log \frac{\exp(\alpha _{y_i}^\top f_i + \beta _{y_i})}{\sum_{j \in \mathcal{C}^{seen}}  \exp (\alpha _{j}^\top f_i + \beta _{j})}, 
    \label{eq:5}  
\end{equation}    
where $\alpha$ and $\beta$ is the weight and bias of the classifier, and $N$ denotes the number of samples in a training batch.

To bridge the seen and unseen categories for zero-shot setting, auxiliary semantic knowledge should be introduced to transfer the knowledge learned from the seen category. Inspired by \cite{liu2019semantic}, we employ a teacher model pre-trained on ImageNet to help our DSN model learn transferrable representations. 

In detail, we utilize the prediction of the teacher model $p \in \mathbb{R}^{|\mathcal{C}^{T}|}$ as the label information instead of one-hot vector used in cross-entropy loss, where $\mathcal{C}^{T}$ denotes the label space of the teacher model and $\mathcal{C}^{T} \cap \mathcal{C}^{unseen}=\varnothing $ in our zero-shot setting. Finally, the auxiliary semantic knowledge (ASK) loss can be formulated as:
\begin{equation}
    \mathcal{L}_{ask}=-\sum_{i = 1}^{N}\sum_{k \in \mathcal{C}_T}p'_{i,k}\log \frac{\exp (\gamma _{k}^\top f_i + \delta _{k})}{\sum_{j \in \mathcal{C}^{T}}  \exp (\gamma _{j}^\top f_i + \delta _{j})},   
\end{equation}  
where $p'$ denotes the teacher signal $p$ with semantic constraints proposed in \cite{liu2019semantic} and $\gamma , \delta$ are weight and bias of the another classifier, respectively.

The total discriminative learning loss can be formulated as:
\begin{equation}
    \mathcal{L}_{dl}= \mathcal{L}_{cls} + \mathcal{L}_{ask}.
\end{equation}

\subsection{Objective and Optimization}
In summary, the full objective function of our DSN model is:
\begin{equation}
    \mathcal{L}= \lambda _1 \mathcal{L}_{cmcm} + \lambda _2 \mathcal{L}_{ml} + \lambda _3 \mathcal{L}_{dl},
\end{equation} 
where $\lambda _1$, $\lambda _2$ and $\lambda _3$ are the hyperparameters to balance the performance of our model.

\begin{table*}[t]
    \begin{center}
        
    \begin{threeparttable}
    \renewcommand\arraystretch{1.1}
    \begin{tabular}{c|c|c|cc|cc}
        \Xhline{0.8pt}
        \multirow{2}*{Task}&\multirow{2}*{Methods}&\multirow{2}*{Dimension}&\multicolumn{2}{c|}{Sketchy}&\multicolumn{2}{c}{TU-Berlin} \\
        \cline{4-7} 
        &&&mAP@all&Prec@100&mAP@all&Prec@100 \\
        \hline
        \multirow{5}*{SBIR}&Siamese CNN \cite{qi2016sketch}&64&0.132&0.175&0.109&0.141 \\
        &SaN \cite{yu2017sketch}&512&0.115&0.125&0.089&0.108 \\
        &GN Triplet \cite{sangkloy2016sketchy}&1024&0.204&0.296&0.175&0.253 \\
        &$\text{DSH}_b$ \cite{liu2017deep}&64&0.171&0.231&0.129&0.189 \\
        &$\text{GDH}_b$ \cite{zhang2018generative}&64&0.187&0.259&0.135&0.212 \\
        \hline
        \multirow{3}*{ZSL}&SAE \cite{kodirov2017semantic}&300&0.216&0.293&0.167&0.221 \\
        &FRWGAN \cite{felix2018multi}&512&0.127&0.169&0.110&0.157 \\
        &$\text{ZSH}_b$ \cite{yang2016zero}&64&0.159&0.214&0.141&0.177 \\
        \hline
        \multirow{11}*{ZS-SBIR}&$\text{ZSIH}_b$ \cite{shen2018zero}&64&0.258&0.342&0.223&0.294 \\
        &CVAE \cite{yelamarthi2018zero}&4096&0.196&0.284&0.005&0.001 \\
        &SEM-PCYC \cite{dutta2019semantically}&64&0.349&0.463&0.297&0.426 \\
        &$\text{SEM-PCYC}_b$ \cite{dutta2019semantically}&64&0.344&0.399&0.293&0.392 \\
        &$\text{SAKE}_b$ \cite{liu2019semantic}&64&0.364&0.487&0.359&0.481 \\
        &CSDB \cite{dutta2019style}&64&0.376&0.484&0.254&0.355 \\
        &DSN (ours)&64&{\bf 0.484}&{\bf 0.610}&{\bf 0.442}&{\bf 0.538} \\
        &$\text{DSN}_b$ (ours)&64&{\bf 0.436}&{\bf 0.553}&{\bf 0.385}&{\bf 0.497} \\
        &SAKE \cite{liu2019semantic}&512&0.547&0.692&0.475&{\bf 0.599} \\
        &DSN (ours)&512&{\bf 0.583}&{\bf 0.704}&{\bf 0.481}&0.586 \\
        &$\text{DSN}_b$ (ours)&512&{\bf 0.581}&{\bf 0.700}&{\bf 0.484}&{\bf 0.591} \\
        \Xhline{0.8pt}
    \end{tabular}
    \end{threeparttable}
    \caption{The retrieval performance comparison with existing SBIR, ZSL and ZS-SBIR methods. (The best performance results are in {\bf bold} and the subscript $b$ denotes binary hash results).}\label{tab:result}
    \end{center}
\end{table*}

\section{Experiments}
\subsection{Datasets}
To verify the effectiveness of our method, We evaluate our DSN method on two popular SBIR datasets, {\it i.e.}, Sketchy \cite{sangkloy2016sketchy} and TU-Berlin \cite{eitz2012hdhso}.
\paragraph{Sketchy.}It consists of 75,471 sketches and 12,500 images from 125 categories. \cite{liu2017deep} extended the image gallery with additional 60,502 photos so there are 73,002 samples in extended Sketchy dataset totally.
\paragraph{TU-Berlin.}It consists of 20,000 sketches uniformly distributed over 250 categories. \cite{liu2017deep} extended this dataset with 204,489 natural images. 

For comparison, we follow the split of \cite{shen2018zero} to randomly choose 25 categories and the rest 100 categories for testing and training in Sketchy. For TU-Berlin, 30 categories and the rest 220 categories are chosen for testing and training. Also,  we follow \cite{shen2018zero} to select at least 400 images for each category in the testing set. 

\subsection{Implementation Details}
We implement our method with PyTorch on two TITAN RTX GPUs. In our model, we utilize a CSE-ResNet-50 pre-trained on ImageNet \cite{lu2018learning} to initialize the backbone. Inspired by \cite{liu2019semantic}, we use the same pre-trained backbone to provide teacher signal as auxiliary semantic knowledge. In the training stage, we train our model with Adam optimizer. The initial learning rate is set to $0.0001$ and exponentially decays to $1e^{-7}$ in training. The batch size in our original setting is 96 and the training epochs is 10. Usually, we set hyperparameters  $\lambda _1 = 0.1$, $\lambda _2 = 1$ and $\lambda _3 = 1$ unless stated. For cross-modal contrastive method, we set the temperature $\tau$ as $0.07$ and the dimension of contrastive vectors is 128 in all the experiments. For category-specific memory bank, we set $k=10$.

\subsection{Comparing with Existing Methods}
We compare our DSN model with various methods related to SBIR, ZSL and ZS-SBIR. Especially for ZS-SBIR, we employ all the experiments under the same setting as others for fairness. In addition, to generate binary hash codes for comparison, we use iterative quantization (ITQ) \cite{gong2012iterative}. Then the hamming distance is computed to conduct retrieval while cosine distance is utilized for real value feature retrieval. The results of our model for ZS-SBIR are shown in Table \ref{tab:result}.  

As we can see, our DSN method outperforms other existing methods especially in Sketchy dataset with 64-d features. Specifically, in 64-d real value feature retrieval, our DSN outperforms the state-of-the-art (SOTA) method \cite{dutta2019style,dutta2019semantically} by 29\% in Sketchy and 49\% in TU-Berlin dataset, respectively. In 512-d real value feature retrieval, our DSN outperforms the SOTA \cite{liu2019semantic} by 7\% in Sketchy and 1\% in TU-Berlin. For binary hash retrieval, we conduct ITQ \cite{gong2012iterative} following the same setting with \cite{liu2019semantic} and outperform SOTA by 16\% and 7\% for Sketchy and TU-Berlin respectively in 64-d setting. Also, we provide 512-d binary hash code retrieval results in the Table \ref{tab:result}, which is similar to 512-d real value feature results. It demonstrates that our DSN method can not only generate accurate real value features but also perform stably even in binary hash setting for ZS-SBIR.  

All of these results show that our DSN method can alleviate the domain gap between sketch and domain effectively, and reduce the large intra-class diversity in sketch domain which is ignored by most existing methods. 

\subsection{Experiment Analysis}
                

\begin{figure}[t]
    \centering
    \subfigure{
    \begin{minipage}[t]{0.5\linewidth}
    \centering
    \includegraphics[width=1.5in]{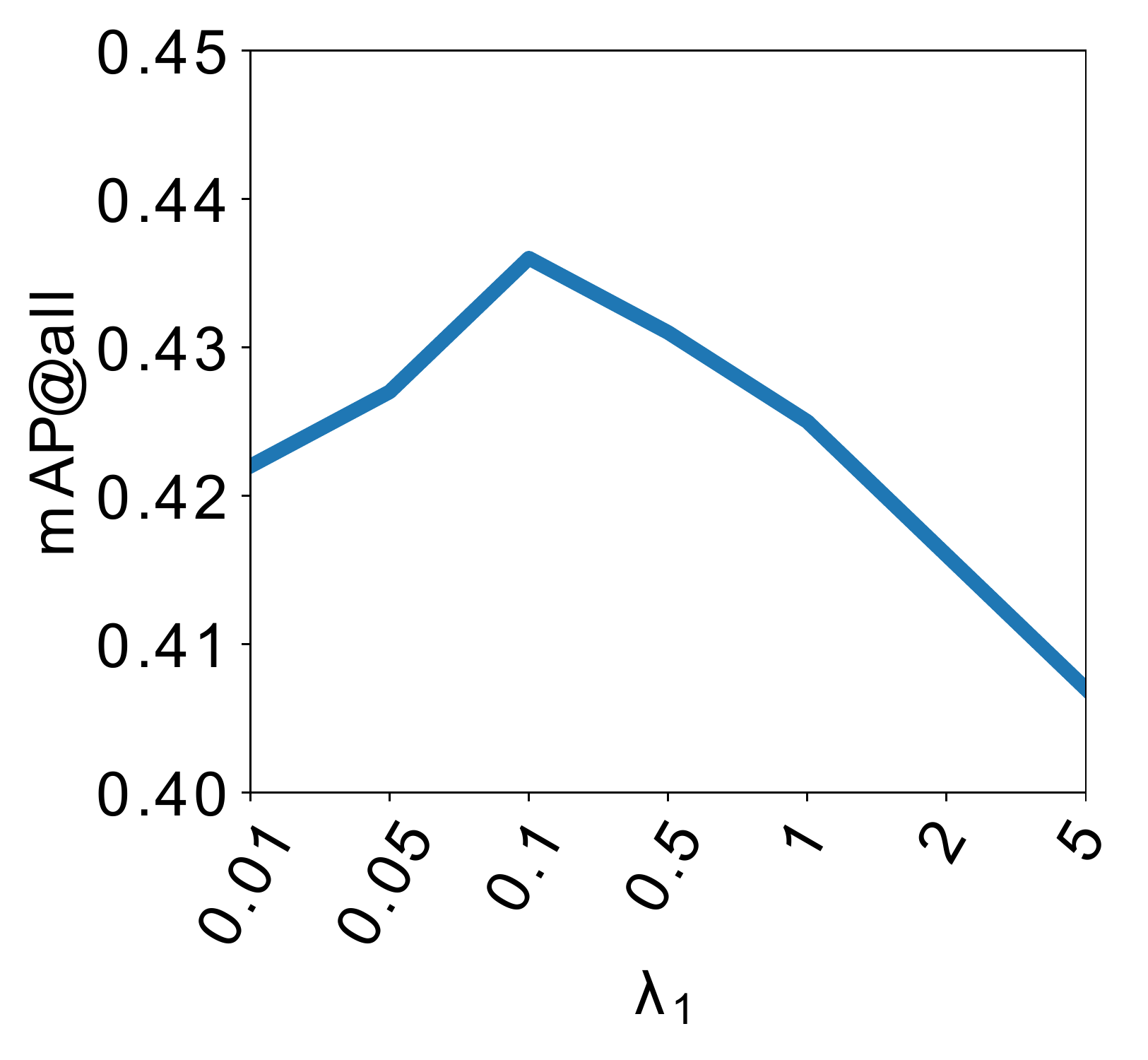}
    \end{minipage}%
    }%
    \subfigure{
    \begin{minipage}[t]{0.5\linewidth}
    \centering
    \includegraphics[width=1.5in]{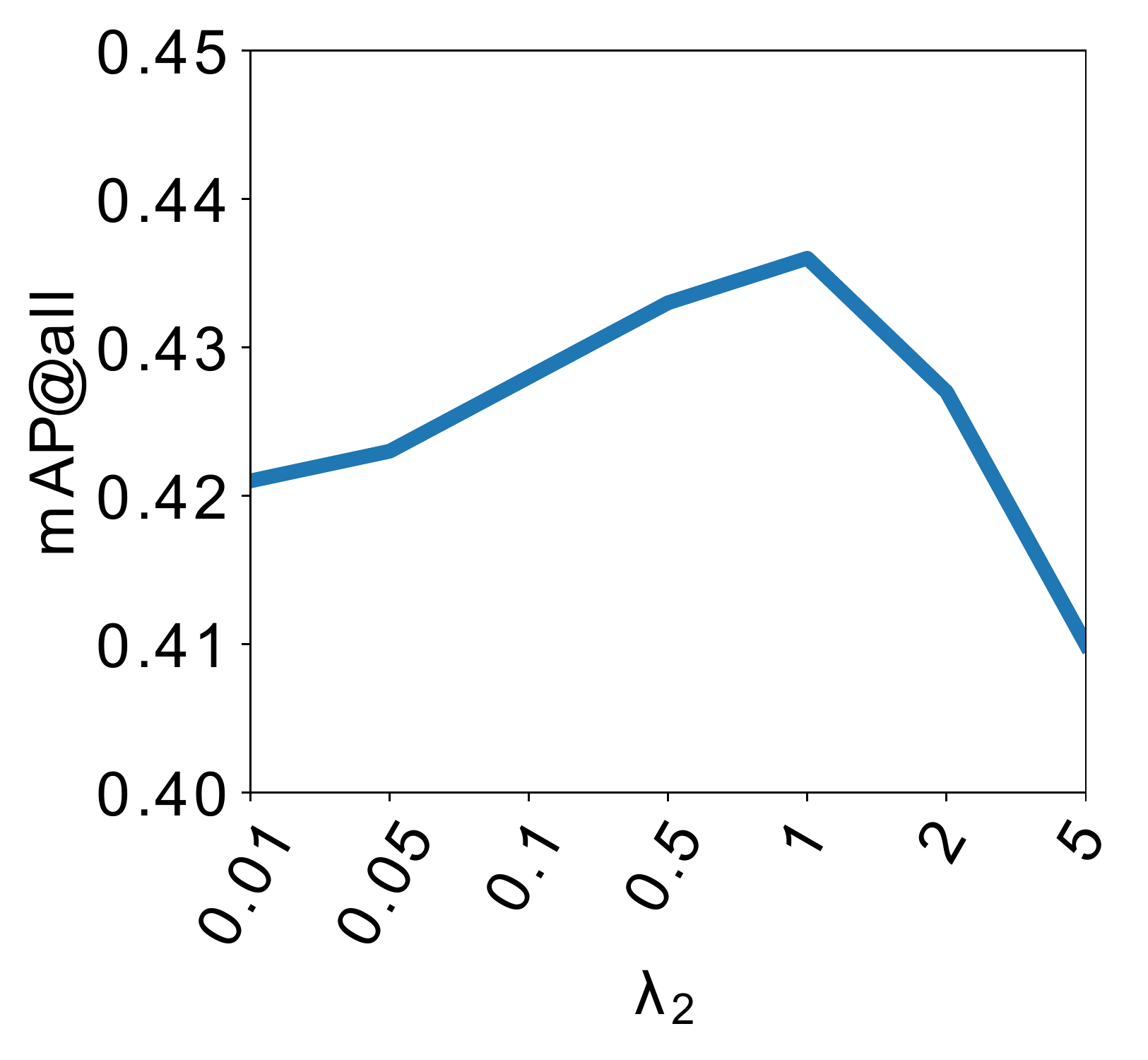}
    \end{minipage}%
    }%
    \vspace{-0.4cm}
    \caption{The 64-d binary hash retrieval results with different $\lambda _1$ for cross-modal contrastive method loss and $\lambda _2$ for memory loss in Sketchy dataset.}
    \centering
    
    \label{fig:quantitative}
\end{figure}

\subsubsection{Quantitative Analysis}

\begin{figure}[t]
    \begin{center}
    \includegraphics[width=1.0\linewidth]{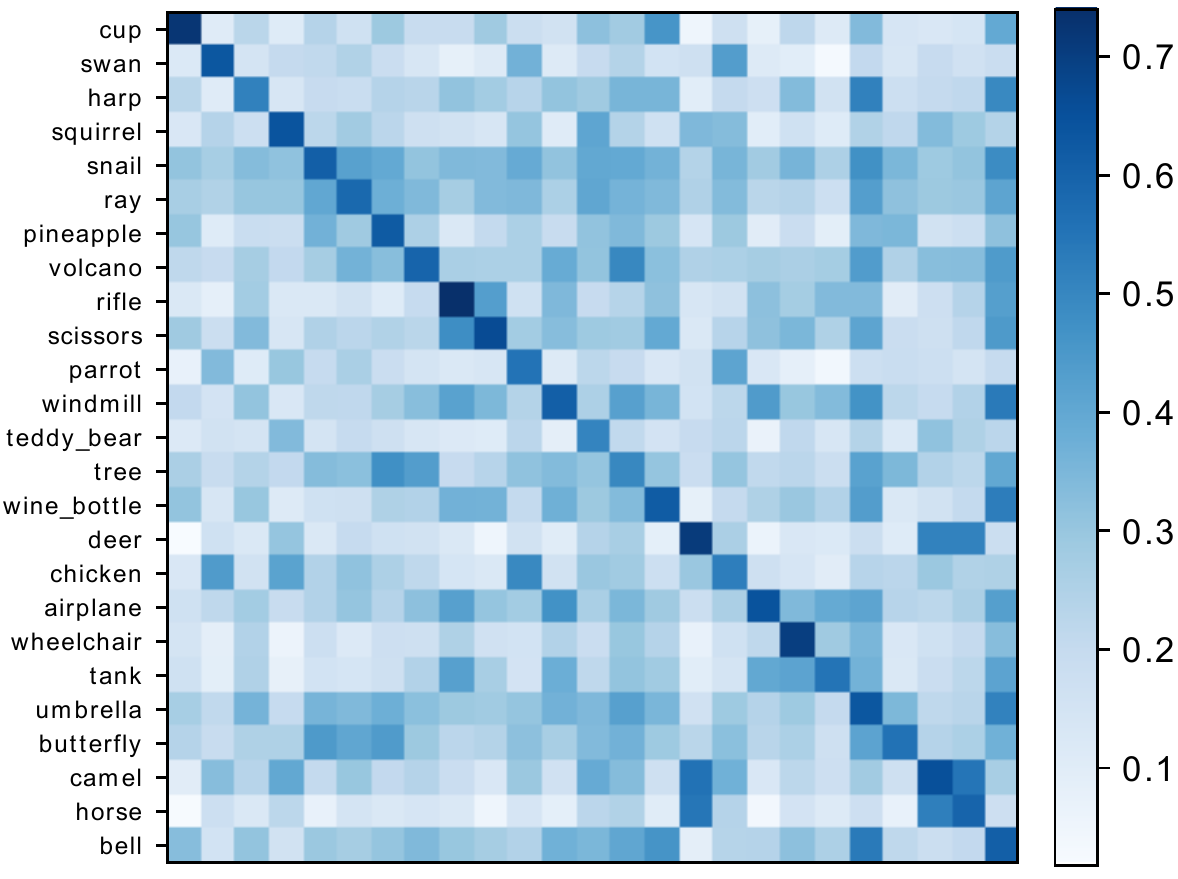}
    \caption{The similarity matrix between the sketches and images with 64-d retrieval features in the testing set of Sketchy dataset. The horizontal and vertical axes represent image and sketch categories respectively. Best view in color.}\label{fig:similarity}
    \end{center}
 \end{figure}

\begin{figure}[t]
    \begin{center}
    \includegraphics[width=0.9\linewidth]{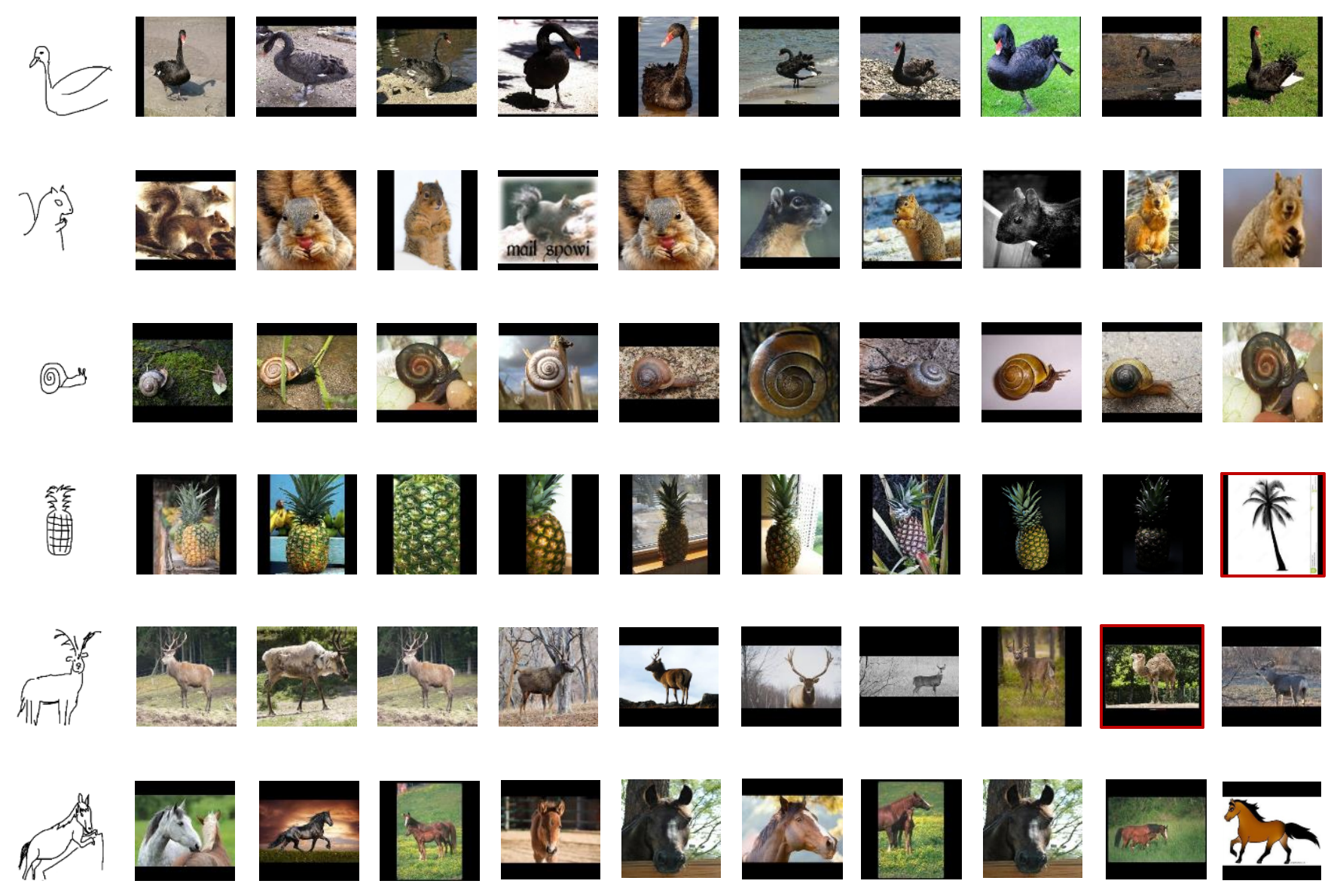}
    \caption{The top 10 zero-shot sketch-based image retrieval results using our DSN model with 512-d real value feature in Sketchy dataset. The mis-retrieved images are highlighted in {\color{red} red} border.}\label{fig:retrieval}
    \end{center}
 \end{figure}

Inspired by \cite{liu2019semantic}, we employ a pre-trained teacher model to provide teacher signal as auxiliary semantic knowledge. So in our experiments, we set $\lambda _3$ to 1 and explore different $\lambda _1$ and $\lambda _2$ for our proposed method.

In Figure \ref{fig:quantitative}, we show mAP@all results with different hyperparameters. As we can see, setting $\lambda _1=0.1$ and $\lambda _2=1$ can achieve the best performance with 64-d binary hash codes in Sketchy. It should be noted that we adopt this setting in all the experiments unless stated otherwise.

\subsubsection{Qualitative Analysis}
\paragraph{Visulization of Similarity.}To demonstrate the superiority of our method, we also visualize the similarity between sketch and image in retrieval feature level (seen Fig. \ref{fig:similarity}). Here, we calculate the cosine similarity using the mean sketch feature and mean image feature for each category in the testing set of Sketchy. The darkness of matrix color represents the similarity. As shown in Figure \ref{fig:similarity}, our DSN method smooths the domain gap to learn discriminative features, which can boost the performance of retrieval. However, negative cases still exist. For example, the camel and horse are similar in appearance, which leads to a higher similarity value in the matrix compared with other pairs of categories. 

\paragraph{Examples of Retrieval.}We show top 10 ZS-SBIR results using our DSN model with 512-d real value features in Figure \ref{fig:retrieval}. The mis-retrieved results are in the red border. As we can see from these examples, our DSN model performs well in most cases except for some visual similarly images. For example, in the retrieval results of pineapple sketch (the 4$_{\text{th}}$ row in Figure \ref{fig:retrieval}), a tree image is mis-retrieved, which is similar to the pineapple query in appearance. 

\subsection{Ablation Study}
\begin{table}
    \begin{center}
        
        \begin{threeparttable}
            \renewcommand\arraystretch{1.1}
            \begin{tabular}{c|c|c}
                \Xhline{0.8pt}
                Models&Sketchy&TU-Berlin \\
                \hline
                \multicolumn{1}{l|}{Baseline}&0.361&0.355\\
                \multicolumn{1}{l|}{Baseline + $\mathcal{L}_{cmcm}$}&0.420&0.373 \\
                \multicolumn{1}{l|}{Baseline + $\mathcal{L}_{ml}$}&0.412&0.372 \\
                \multicolumn{1}{l|}{Baseline + $\mathcal{L}_{cmcm}$ + $\mathcal{L}_{ml}$}&{\bf 0.436}&{\bf 0.385} \\
                \Xhline{0.8pt}
                
            \end{tabular}
        \end{threeparttable}
        \caption{The mAP@all results with 64-d binary hash codes of our ablation study. (The best results are in {\bf bold}).}\label{tab:ablation}
    \end{center}    
\end{table}

We conduct four ablation studies to validate the performance of our DSN model: 1) A baseline that learns a common representation for sketches and images with a pre-trained model as teacher model inspired by \cite{liu2019semantic}; 2) Adding our cross-modal contrastive method $\mathcal{L}_{cmcm}$ to the baseline that smooths the domain gap by mining relations between samples; 3) Adding our category-specific memory bank $\mathcal{L}_{ml}$ to the baseline that reduces the large intra-class diversity; 4) Full DSN model. The experiments are conducted with 64-d binary hash codes for retrieval in Sketchy dataset. 

As shown in Table \ref{tab:ablation}, $\mathcal{L}_{cmcm}$ can improve baseline performance by 16\% and 5\% in Sketchy and TU-berlin respectively, while $\mathcal{L}_{ml}$ can improve baseline performance by 14\% and 5\%. However, combining $\mathcal{L}_{cmcm}$ with $\mathcal{L}_{ml}$ improves the baseline only by 21\% and 8\% for each dataset. It should be noted that the improvement with $\mathcal{L}_{cmcm} + \mathcal{L}_{ml}$ isn't a simple addition operation with $\mathcal{L}_{cmcm}$ and $\mathcal{L}_{ml}$. In our knowledge, the reason may be that our cross-modal contrastive method has already reduced the intra-class diversity in sketch domain to some extent, which pulls the sketches with a same label together in a training batch. These two novel methods have intersection in reducing intra-class diversity, which results in the phenomenon mentioned above. Besides, the experiment results demonstrating $\text{Imp}(\mathcal{L}_{cmcm}) > \text{Imp}(\mathcal{L}_{ml})$, where Imp denotes the improvement from baseline, confirms our conjecture from another perspective.

\section{Conclusion}
In this paper, we present a novel yet effective Domain-Smoothing Network for ZS-SBIR. First, a cross-modal contrastive method is proposed to smooth the domain gap by mining relations between sketches and images with sufficient positive and negative samples. Second, we build a category-specific memory bank to address the large intra-class diversity in the sketch domain, where prototypes of each sketch category are learned to reduce the variability in appearance. We conduct extensive experiments on two popular datasets, {\it i.e.}, Sketchy and TU-Berlin, which demonstrate our DSN method outperforms existing methods by a large margin and validate its superiority.

\section*{Acknowledgments}
	
Our work was supported in part by the National Natural Science Foundation of China under Grant 62071361, and in part by the Fundamental Research Funds for the Central Universities ZDRC2102.

\clearpage

\bibliographystyle{named}
\bibliography{ijcai21}

\end{document}